\documentclass[acmsmall]{acmart}

\usepackage{hyperref}
\usepackage{natbib}
\usepackage{amsmath,amsfonts}
\usepackage{cases}
\usepackage{algorithmic}
\usepackage{graphicx}
\usepackage{textcomp}
\usepackage{xcolor}
\usepackage{booktabs}
\usepackage{colortbl}

\usepackage[linesnumbered,ruled,vlined]{algorithm2e}
\SetKw{Continue}{continue}
\SetKw{Or}{or}
\usepackage{enumitem}
\usepackage{tikz}
\newcommand*\circled[1]{\tikz[baseline=(char.base)]{\node[shape=circle,draw,inner sep=1pt] (char) {#1};}}
\DontPrintSemicolon
\usepackage{multirow}
\usepackage{nicefrac}
\usepackage{tabularx}
\usepackage{mathtools}
\usepackage{pifont}

\newcommand{\xmark}{\ding{55}}
\usepackage{siunitx}
\usepackage{subcaption}
\usepackage{microtype}
\usepackage{makecell}
\usepackage{listings}
\usepackage{color}

\newcommand*{\belowrulesepcolor}[1]{%
	\noalign{%
		\kern-\belowrulesep
		\begingroup
		\color{#1}%
		\hrule height\belowrulesep 
		\endgroup
	}%
}
\newcommand*{\aboverulesepcolor}[1]{%
	\noalign{%
		\begingroup
		\color{#1}%
		\hrule height\aboverulesep
		\endgroup
		\kern-\aboverulesep
	}%
}

\lstdefinestyle{customc}{
  belowcaptionskip=1\baselineskip,
  breaklines=true,
  frame=L,
  xleftmargin=\parindent,
  language=C,
  showstringspaces=false,
  basicstyle=\footnotesize\ttfamily,
  keywordstyle=\bfseries\color{green!40!black},
  commentstyle=\itshape\color{purple!40!black},
  identifierstyle=\color{black},
  stringstyle=\color{orange},
}

\lstdefinestyle{customasm}{
  belowcaptionskip=1\baselineskip,
  frame=L,
  xleftmargin=\parindent,
  language=[x86masm]Assembler,
  basicstyle=\footnotesize\ttfamily,
  commentstyle=\itshape\color{purple!40!black},
}

\AtBeginDocument{%
  \providecommand\BibTeX{{%
    \normalfont B\kern-0.5em{\scshape i\kern-0.25em b}\kern-0.8em\TeX}}}

\setcopyright{acmcopyright}
\copyrightyear{2021}
\acmYear{2021}
\acmDOI{10.1145/nnnnnnn.nnnnnnnn}

\acmJournal{TRETS}
\acmVolume{1}
\acmNumber{1}
\acmArticle{0}
\acmMonth{10}

\begin{document}

\title{Design of Distributed Reconfigurable Robotics Systems with ReconROS}

\author{Christian Lienen}
\email{christian.lienen@upb.de}
\orcid{1234-5678-9012}
\author{Marco Platzner}
\email{platzner@upb.de}
\affiliation{%
  \institution{Paderborn University, Computer Science Department}
  \streetaddress{Warburger Str. 100}
  \city{Paderborn}
  \country{Germany}
  \postcode{33098}
}

\email{larst@affiliation.org}

\renewcommand{\shortauthors}{Lienen and Platzner}

\begin{abstract}
Robotics applications process large amounts of data in real-time and require compute platforms that provide high performance and energy-efficiency. FPGAs are well-suited for many of these applications, but there is a reluctance in the robotics community to use hardware acceleration due to increased design complexity and a lack of consistent programming models across the software/hardware boundary.
In this paper we present {\sc ReconROS}, a framework that integrates the widely-used robot operating system (ROS) with ReconOS, which features multithreaded programming of hardware and software threads for reconfigurable computers. This unique combination gives ROS 2 developers the flexibility to transparently accelerate parts of their robotics applications in hardware. We elaborate on the architecture and the design flow for {\sc ReconROS} and report on a set of experiments that underline the feasibility and flexibility of our approach.
\end{abstract}



\maketitle

\section{Introduction}
\label{sec:Introduction}

Robotics systems are often distributed and can involve challenging computational tasks. Resource-efficiency is a fundamental challenge of such systems since large amounts of data must be processed with soft or even hard real-time constraints~\cite{Yanmaz+18}. Compared to implementations on CPUs and GPUs, FPGAs have been shown to offer higher performance and higher energy-efficiency for many of the involved tasks, e.g., for vision kernels~\cite{8782524}, for morphological image processing functions~\cite{7298356}, for feature detection and description algorithms~\cite{7577310}, and for convolutional neural network inference~\cite{8401525}. However, despite the demonstrated advantages of FPGAs, their uptake into the robotics domain is still limited for several reasons. On one hand, FPGA design and, all the more, software/hardware co-design are arguably more challenging than embedded software development. On the other hand, robotics engineers and application developers are typically not trained in FPGA circuit or hardware/software co-design.

High level synthesis (HLS) tools are available today that accept standard C/C++ for describing behavior and (semi-)automatically translate such descriptions to FPGA hardware. Although HLS tools increase productivity and are thus highly useful, a consistent programming model for implementing software and hardware functions is still lacking. Porting a robotics application from software to hardware or accelerating parts of the application in hardware requires the creation of suitable interfaces between software and FPGA hardware and very often leads to a re-development of substantial parts of the application.

Besides the demand for increased computational capacity and robustness, also the distributedness of typical robotics systems caused by the functional decomposition of larger applications into distinct processes that run in parallel, and often under real-time constraints, as well as the potential spatial distribution of these processes pose opportunities and challenges for hardware acceleration. 
There have been approaches for including reconfigurable hardware into distributed embedded systems, for example ReCoNets~\cite{reconet}, LMGS~\cite{lmgs} or RSS~\cite{rss}, but these approaches are not compatible with existing and widely-used software abstractions for creating distributed robotics systems.
 
In our work, we take up a very popular programming environment in the robotics domain, the robot operating system (ROS). ROS is a middleware layer that models applications as set of communicating nodes and provides several communication mechanisms for information exchange. 

In this paper, which is an extension of our previous conference publication~\cite{LP20}, we present the open source project {\sc ReconROS} as a novel integration of ROS with ReconOS~\cite{6636314}. This paper extends the conference publication mainly by more detailed explanations of the design flow and tool chain, by supporting also the ROS 2 communication paradigms actions and services, and by presenting and quantitatively evaluating {\sc ReconROS} on more advanced robotics applications, especially a distributed real-world mechatronics model with multiple FPGAs.
ReconOS provides an architecture and programming model to enable shared memory multi-threading for software and hardware threads. As a result, {\sc ReconROS} allows robotics developers to utilize hardware acceleration for ROS applications either as hardware-accelerated ROS nodes or as ROS nodes mapped completely to hardware. The latter option provides a consistent programming model for ROS applications, independently of the mapping of ROS nodes to software or hardware.

The remainder of the paper is organized as follows: Section~\ref{sec:BackgroundRelatedWork} provides an overview over ROS and related approaches for integrating hardware accelerators into ROS. Section~\ref{sec:DesignConsiderations} elaborates on different approaches for accelerating ROS applications, before Section~\ref{sec:ReconROS} details {\sc ReconROS} with its architecture and design flow. In Section~\ref{sec:Evaluation}, we present experiments to quantify overheads involved when mapping ROS nodes to hardware and to demonstrate the feasibility and flexibility of {\sc ReconROS}. Finally, Section~\ref{sec:ConclusionFuture Work} concludes the paper and gives an outlook to future work.

\section{Background and Related Work}
\label{sec:BackgroundRelatedWork}

In this section, we first briefly introduce the robot operating system (ROS) and then analyze and compare related approaches for integrating FPGA hardware acceleration into ROS.

\subsection{The Robot Operating System (ROS)}
\label{sec:BackgroundRelatedWork:ROS2}
The Robot Operating System (ROS)\footnote{\url{https://www.ros.org}} is an open source middleware on top of Linux for robotics applications that was originally developed by Willow Garage and is now coordinated by the Open Robotic Foundation. ROS comprises a multitude of libraries and an infrastructure for building and reusing robot-related software modules. The ROS programming paradigm splits larger software architectures into nodes, which use certain communication mechanisms for information exchange. 

The decomposition into nodes promises code reusability and modularity for robot architectures. Available communication mechanisms comprise (i) a many-to-many publish/subscribe model, which allows to broadcast messages to multiple subscribers but is one-way, (ii) services that follow a client-server model where the server provides data only if requested by the client, basically mimicking a remote procedure call, and (iii) actions. 
Actions are the most elaborated communication mechanism where a client inquires about a functionality at a server, starts the functionality if it is available, and then receives regular feedback about the server's progress. Technically, actions are implemented in two phases. The first phase corresponds to a ROS service and the second phase comprises a second ROS service and a feedback channel using the ROS publish/subscribe mechanism.

ROS 2 is the latest release of ROS. In earlier versions only one ROS node per Linux process was supported. This prevented the use of shared memory communication when two or more ROS nodes are mapped to the same compute node. While this limitation was mitigated through the support of so-called nodelets, with ROS 2 multiple ROS nodes can natively run within one Linux process and there is support for shared memory communication. ROS 2 is built on top of an exchangeable communication layer, the data distribution service (DDS). DDS is an industry standard for decentralized communication and available from different vendors. Compared to older ROS versions, the use of DDS provides better configurability and improves properties such as scalability, reliability and durability~\cite{ROS 2ddsperformance}.

Another important element of ROS are ROS messages, which are multi-layered combinations of built-in data types such as integers, floats and strings. Besides predefined message types, e.g., for images or 3D point clouds, custom messages can be created. Since the length of a message might vary during runtime, the ROS 2 middleware supports dynamic memory allocation for messages.

\subsection{Related Approaches for ROS-FPGA Integration}
\label{sec:BackgroundRelatedWork:relatedwork}
In the last years, a few approaches have been presented that integrate reconfigurable hardware accelerators into a ROS software architecture.  
Yamashina et al.~\cite{Li2015} proposed so-called ROS-compliant FPGA components. A ROS node is implemented in software and accesses the hardware component, i.e., the accelerator, via a software wrapper. Communication within the ROS network is completely handled in software and, whenever acceleration is needed, only the payload of the ROS message is transmitted to the hardware component. Semantically, the communication between the ROS software wrapper and the hardware accelerator is a remote procedure call, realized in Xilinux. In~\cite{Yamashina2016}, the automated design tool cReComp (creator for reconfigurable component) is presented to help generate ROS-compliant FPGA components and thus reduce development costs. For the implementation of a ROS-compliant FPGA component with cReComp, the developer has to modify a configuration file and create user logic for the hardware accelerator. The configuration file contains information about the interface between the processing system and the programmable logic. cReComp generates the software and hardware parts for this interface. An evaluation by a group of test developers confirmed higher design productivity compared to manually designed interfaces. 

In follow-up work, Sugata et al.~\cite{Sugata2017} identify the communication times between ROS nodes as bottlenecks and aim to reduce these times through implementing the ROS publish/subscribe messaging in hardware. In their system, communication is divided into two phases: the connection establish phase, which is supported by software, and the data communication phase that is realized by two network stacks implemented in FPGA hardware. This  reduces the communication time between nodes by 50 percent.
Ohkawa et al.~\cite{8823798} extend this work by using high level synthesis (HLS) for accelerator implementation and ROS protocol interpretation to increase productivity. Their approach takes the ROS message definition, the ROS node configuration, and behavioral code written in C/C++ for the accelerator and generates the FPGA design. The infrastructure of the generated design includes several components: the hardwired TCP/IP stacks for the data communication phase, a data conversion between ROS messages and the application, an interface between the data conversion and the application, and, finally, the application itself.  

Leal et al.~\cite{leal2020automated} present Forest, an approach for combining the more recent release ROS 2 with hardware acceleration. Forest uses configuration files to specify so-called ROS 2-FPGA nodes, which are a composition of a ROS 2 software node, an HLS-coded FPGA hardware module, and a PYNQ driver for the interaction between the ROS 2 software node and the hardware module. 

While~\cite{Sugata2017,8823798} migrate almost a complete ROS node to hardware, Podlubne and G\"ohringer~\cite{Podlubne2020} go one step further and propose a methodology for full-hardware implementation of a number of ROS nodes. Their hardware designs comprise four parts: the ROS application nodes that use publish/subscribe communication, a so-called application-to-ROS converter, a communication interface, and a manager. Basically, the application-to-ROS converter serializes the ROS-based IP traffic on an AXI bus, the communication interface handles the AXI messages and sends them to a TCP/IP stack to connect to external ROS nodes, and the manager coordinates the communication between the ROS nodes and the TCP/IP stack. 
Conceptually, the application-to-ROS converter must reside in hardware, but the communication interface and the manger could also be mapped to the processing system of the platform FPGA. 
However, the main feature of this methodology is the option to implement one or more ROS nodes fully in hardware and map them to reconfigurable logic without the need of using a processor. Likewise, any application implemented in reconfigurable hardware can be made ROS-compatible. Furthermore, the presented implementation can use dynamic custom ROS messages.

Strohmer et al.~\cite{8994770} presented a ROS-enabled hardware framework for experimental robotics. They use the programmable logic on a Xilinx Zynq-7000 for signal conditioning and partition the available CPU cores into a non real-time part running Linux with ROS and a real-time part running control algorithms. A distributed network of FPGAs can extend the signal conditioning part using TosNet, which provides memory access across multiple nodes by memory mirroring.

Eisoldt et al.~\cite{eisoldt2021reconfros} contributed ReconfROS, a framework for ROS hardware acceleration based on shared-memory communication. The architecture on the system-on-chip comprises a software part including a ROS node, a shared memory area, and one or more processing blocks in the programmable logic. The software-mapped ROS node subscribes to topics and writes received messages into the shared memory area, from where the data can be accessed by the hardware processing blocks. Finally, the software-mapped ROS node publishes the resulting data. The control of the processing blocks is done via control registers which are mapped into the virtual address space of the software application.

\section{Design Considerations}
\label{sec:DesignConsiderations}

\begin{figure}[thb]
    \includegraphics[width=0.7\linewidth]{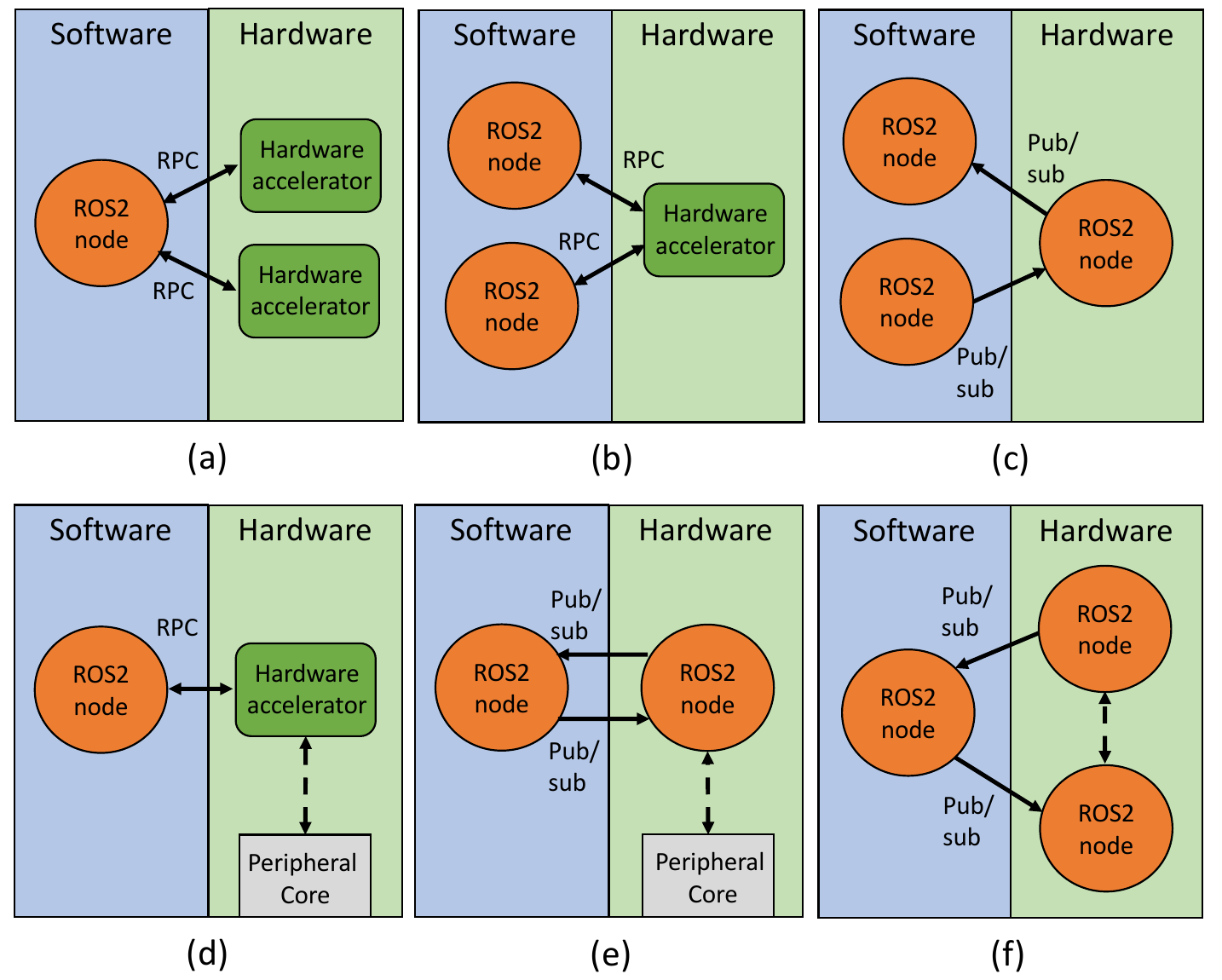}    
    \caption{Different schemes for integrating ROS 2 node with hardware accelerators}
    \label{fig:design_considerations}
\end{figure}

The goal of this work is to provide developers of ROS 2-based robotics applications with a flexible means to utilize programmable logic for hardware acceleration. On the level of ROS 2 applications, there are several schemes for such an integration, which are sketched in Figure~\ref{fig:design_considerations}. Figure~\ref{fig:design_considerations}(a) shows a scheme where some parts of a ROS 2 node, typically runtime-consuming functions, are mapped to one or several accelerators in programmable logic. The semantics of the communication between the ROS 2 node and the accelerators is that of a remote procedure call (RPC). In Figure~\ref{fig:design_considerations}(b), a hardware accelerator is shared between several ROS 2 nodes. Communication semantics is still RPC, but the implementation is more involved since proper arbitration between the accesses of the ROS 2 nodes is required. The third scheme shown in Figure~\ref{fig:design_considerations}(c) is the most advanced and allows to map complete ROS 2 nodes to hardware. Essentially, the hardware accelerator is turned into a ROS 2 node. In this scheme, all ROS 2 nodes can communicate via the ROS 2 communication mechanisms, independently of their mapping to software or hardware. Semantically, this is the most intriguing scheme since it provides a consistent programming model across hardware and software where all ROS 2 nodes use exactly the same ROS 2 functions. 
 
Often, developers decide to attach interfaces to sensors and actuators directly to the reconfigurable hardware and provide peripheral cores in hardware to access them rather than putting them under operating system control on the host CPU. Figure~\ref{fig:design_considerations}(d) and Figure~\ref{fig:design_considerations}(e) sketch such schemes with dashed lines. While these schemes are popular for maximizing performance in concrete robotics applications, there are also two possible pitfalls: 
First, flexibility is reduced since directly connected peripherals can not be accessed by other ROS 2 nodes, and much less so when the ROS 2 nodes are mapped to different compute nodes in a distributed system. Second, many sensors and actuators come with standardized interfaces and corresponding drivers, e.g., USB, for which the use of an existing, software-accessible peripheral of the compute platform is much more productive than to implement suitable interfaces and protocol stacks in hardware. 
Along the same line, the scheme shown in Figure~\ref{fig:design_considerations}(f) directly connects several ROS 2 nodes mapped to hardware without relying on ROS 2 communication mechanisms. This can increase performance in particular cases, but again lacks flexibility since the mapping of the ROS 2 nodes is severely constrained.

{\sc ReconROS}\footnote{https://github.com/Lien182/ReconROS} integrates the ROS 2 middleware with the ReconOS/Linux architecture and programming model for hardware/software multithreading on platform FPGAs and can realize all schemes shown in Figure~\ref{fig:design_considerations}(a)-(f) and their combinations. 
On one hand, ReconOS enables us to develop applications as a set of software and hardware threads under the shared memory model. On the other hand, ROS 2 allows for declaring several ROS 2 nodes within one Linux process. Therefore, in the schemes shown in Figure~\ref{fig:design_considerations}(a)(b)(d) each hardware accelerator is encapsulated by a ReconOS hardware thread. In contrast to most of related work, {\sc ReconROS} hardware accelerators can communicate with the ROS 2 software nodes not only by passing data in an RPC manner, but can also use shared memory communication in the Linux virtual address space, which is more efficient when larger data structures have to be passed. In such a case, pointers to arbitrarily large ROS 2 messages are passed and the accelerators themselves retrieve the relevant message payload from shared memory. Furthermore, since ReconOS hardware threads can execute standard operating system synchronization primitives, the required arbitration for the scheme in Figure~\ref{fig:design_considerations}(b) is straight-forward to realize. 
In the more advanced schemes shown in Figure~\ref{fig:design_considerations}(c)(e)(f), ReconOS hardware threads implement complete ROS 2 nodes and allow them to access operating system functions and also ROS 2 communication primitives, using the whole set of standard and even custom-defined ROS messages.

Table~\ref{table:BackgroundRelatedWork:comparison} compares {\sc ReconROS} with related approaches. In contrast to all other approaches except for \cite{leal2020automated}, {\sc ReconROS} leverages the more future-oriented ROS 2 version which promises improved scalability and real-time properties. Hardware acceleration of a ROS node mostly implies to partition the node and implement it as hardware/software co-design. This is followed by all approaches except~\cite{Podlubne2020}. %
Mapping several ROS nodes to hardware is possible in~\cite{eisoldt2021reconfros} and {\sc ReconROS}. Full memory access for hardware accelerators and arbitrarily long ROS messages are featured by~\cite{eisoldt2021reconfros} and {\sc ReconROS}. A consistent hardware/software programming model and the support of all available ROS 2 communication paradigms are unique features of {\sc ReconROS}. 

\begin{table}
    \begin{center}
      \begin{tabular}{|l c c c c c c|} 
        \hline
        Characteristic & \makecell{\cite{Li2015},\cite{Yamashina2016}, \\ \cite{Sugata2017},\cite{8823798} }    & \cite{Podlubne2020}   & \cite{8994770} & \cite{eisoldt2021reconfros} & 
		\cite{leal2020automated}   & \makecell{{\sc ReconROS}} \\ [0.5ex] 
        \hline
        \makecell[l]{ROS version}            & 1                         & 1                     & 1   &   1    & 2       & 2 \\ 
         \hline
         \makecell[l]{Support of hardware/software \\ co-designed ROS nodes}    & \checkmark                & \xmark         & \checkmark       & \checkmark & \checkmark        & \checkmark \\
         \hline
	     \makecell[l]{Multiple ROS \\ nodes per FPGA}                          & \xmark                    & \checkmark            & \xmark     & \xmark    & \xmark   & \checkmark \\
 	   \hline
       \makecell[l]{Consistent hardware/software \\ programming model}                   & \xmark                    & \xmark                & \xmark  & \xmark    & \xmark       & \checkmark \\
         \hline
         \makecell[l]{Memory access \\ for hardware accelerators}                    & \xmark                    & \xmark                & \xmark & \checkmark & \xmark          & \checkmark \\
		 \hline
	    \makecell[l]{Support of arbitrarily  \\ long ROS messages}                   & \xmark                    & \xmark                & \xmark       & \checkmark  & \xmark   & \checkmark \\
   \hline
        \makecell[l]{Support of ROS \\ services and actions}                   & \xmark                    & \xmark  & \xmark        & \xmark      & \xmark            & \checkmark \\

	 \hline
       \end{tabular}
     \end{center}
    \caption{Comparison of approaches for integrating hardware accelerators with ROS}
    \label{table:BackgroundRelatedWork:comparison}
\end{table}

\section{{\bf \sc ReconROS}}
\label{sec:ReconROS}

In this section, we present the architecture of {\sc ReconROS}, followed by the design flow and an example that shows the programming interface.

\begin{figure*}[ht]
	\centering
    \includegraphics[width=\textwidth]{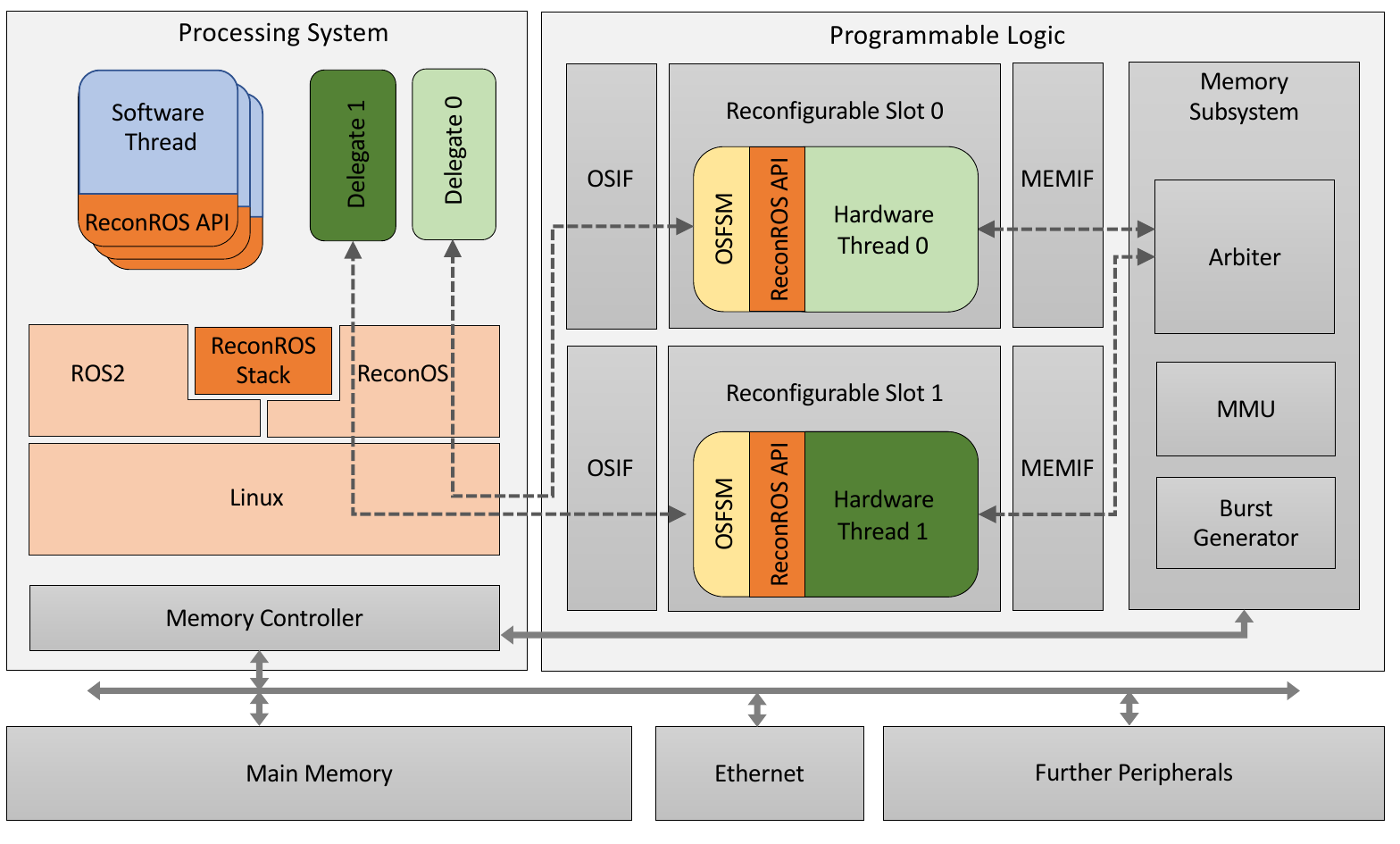}    
    \caption{{\sc ReconROS} architecture with two hardware ROS 2 nodes (threads) and several software ROS 2 nodes (threads)}
    \label{fig:reconros_architecture}
\end{figure*}

\subsection{Hardware/Software Architecture}
\label{sec:ReconROS:architecture}

{\sc ReconROS} inherits most of its hardware architecture from the underlying ReconOS~\cite{6636314,Luebbers_Platzner_2009}. Figure~\ref{fig:reconros_architecture} shows an example architecture with two hardware ROS 2 nodes (threads) and several software ROS 2 nodes (threads). The hardware threads are mapped to reconfigurable slots and are connected to the Linux operating system kernel running on the CPU via the operating system interface (OSIF) and to shared memory via the memory interface (MEMIF). A so-called operating system finite state machine (OSFSM) is attached to each hardware thread to serialize the thread's operating system interactions. On the CPU, the communication with the OSIF is handled by a ReconROS driver and by light-weight delegate threads that serve the operating system calls for the hardware threads. The memory subsystem enables the hardware threads to access the whole address space of the {\sc ReconROS} application, including shared memory and memory-mapped peripherals. ReconOS supports virtual memory and therefore includes an MMU in its memory subsystem.

\begin{table}[]
    \begin{tabular}{|lll|}
    \hline
    ReconROS object   & ROS 2 equivalent         & Description                                                                                                       \\
    \hline
    rosnode           & node                    & \begin{tabular}[c]{@{}l@{}}Represents a ROS 2 node (software or hardware) \\ in the  ReconROS Stack\end{tabular}                            \\
    \hline
    rosmsg            & message                 & \begin{tabular}[c]{@{}l@{}}Message type for communication mechanisms  \\ publish/subscribe, service, or action\end{tabular}           \\
    \hline
    rossub            & subscriber              & \begin{tabular}[c]{@{}l@{}}Enables a rosnode to subscribe  to a topic \\ using a 
specific rosmsg\end{tabular}  \\
    \hline
    rospub            & publisher               & \begin{tabular}[c]{@{}l@{}}Enables a rosnode to publish to a topic \\ using a
specific rosmsg\end{tabular}            \\
    \hline
    rossrvs / rossrvc  & service server / client & \begin{tabular}[c]{@{}l@{}}Extends a rosnode by the capability to act as \\ server or client for ROS 2 services\end{tabular} \\
    \hline
   rosacts / rosactc & action server / client  &  \begin{tabular}[c]{@{}l@{}}Extends a rosnode by the capability to act as \\ server or client for ROS 2 actions\end{tabular}  \\
    \hline    
    \end{tabular}
    \caption{Objects of the {\sc ReconROS} stack}
    \label{table:reconros:primitives}
\end{table}

To realize {\sc ReconROS}, we have developed two additional components, (i) the {\sc ReconROS} stack and (ii) the {\sc ReconROS} API for software and hardware threads. The {\sc ReconROS} stack extends the existing set of ReconOS objects such as semaphores or mailboxes with ROS 2-related objects. Table~\ref{table:reconros:primitives} lists the objects of the {\sc ReconROS} stack. ROS 2 nodes mapped to either software or hardware can create these objects and call corresponding methods in exactly the same way.

The {\sc ReconROS} API abstracts the standard ROS 2 API and allows ReconOS threads to access the objects of the {\sc ReconROS} stack. As indicated in Figure~\ref{fig:reconros_architecture}, the {\sc ReconROS} API is available for both software and hardware threads. While due to the flexibility of the underlying ReconOS system any ROS 2 function can be made available for hardware threads, the current set of provided functions dealing with the objects listed in Table~\ref{table:reconros:primitives} is sufficient to implement ROS 2 hardware nodes that receive data, process it, and send it back. In particular, ROS 2 hardware nodes can publish and subscribe to topics and assume both server and client roles in ROS 2 services and actions. Software threads can not only access the {\sc ReconROS} API but also the standard ROS 2 API to utilize a richer set of functions.

In contrast to most of related work, our ROS 2 hardware nodes can access shared memory and thus implement a more efficient ROS message handling. When hardware threads access functions of the {\sc ReconROS} API, e.g., for subscribing or publishing to topics, the OSIF and the delegate thread mechanism are used to pass pointers between the {\sc ReconROS} stack in software and the hardware threads to allow them to access the ROS message data structures in memory through their MEMIFs. Compared to message communication via the OSIF, which corresponds roughly to the mechanism used in most of related work, this design decision brings about two advantages: First, the MEMIF interface provides higher data rates due to the used AXI high performance interface of the processing system. Second, the transmission of the data can be done without using the processing system, which leads to more potential for parallel execution of software and hardware threads.

\begin{figure}[ht]
    \includegraphics[width=0.7\linewidth]{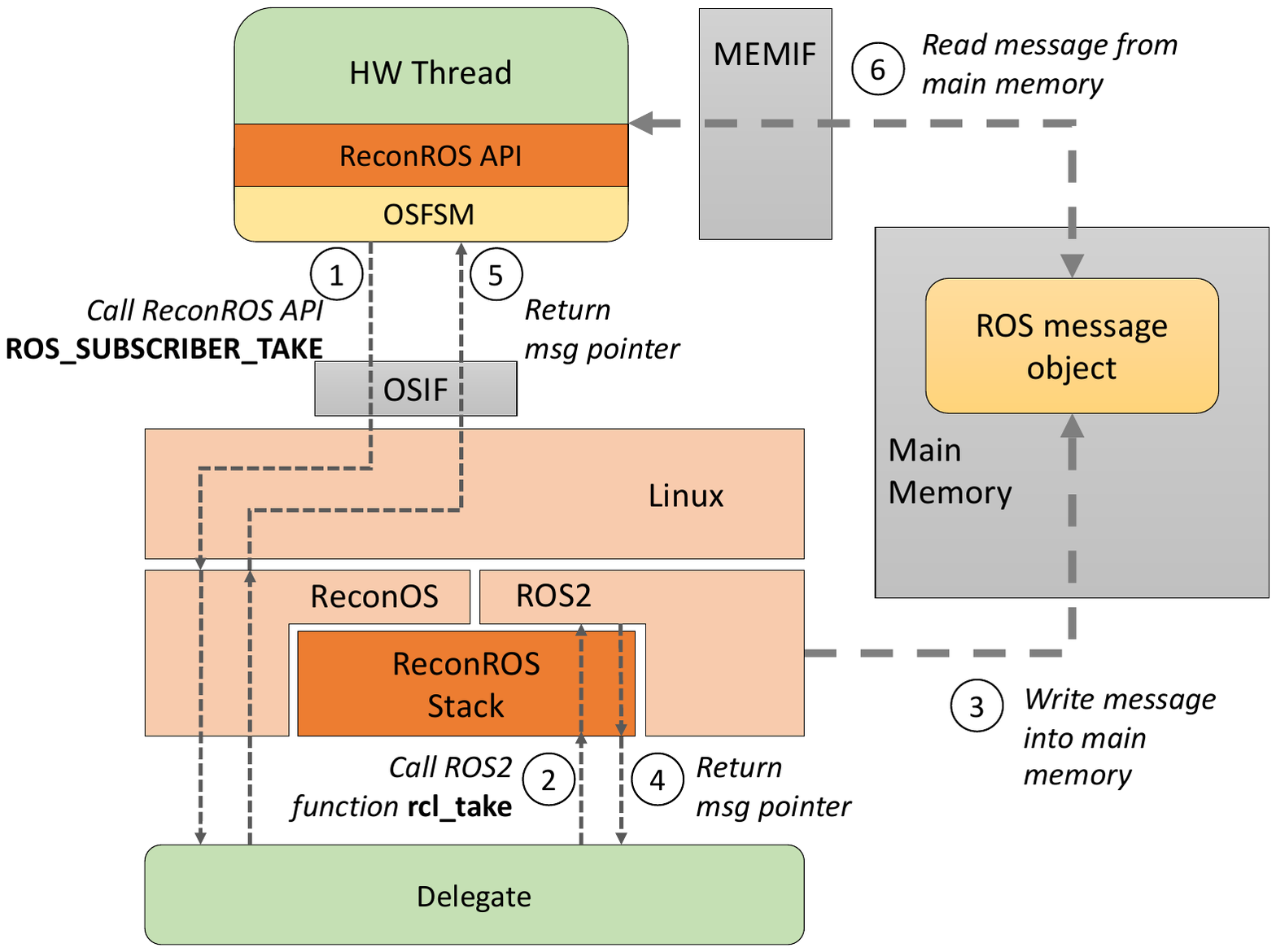}      
    \caption{Sequence of events when a ROS 2 hardware node calls the \textit{ROS\_SUBSCRIBER\_TAKE} function from the {\sc ReconROS} API}
    \label{fig:reconros_take_sequence}
\end{figure}

Figure~\ref{fig:reconros_take_sequence} exemplifies the sequence of events when a hardware ROS 2 node initiates the function {\tt ROS\_SUBSCRIBER\_TAKE} from the {\sc ReconROS} API \circled{1}. The function call of the hardware thread includes the command for this API function and a reference to the subscriber. The command is transmitted by the OSFSM and unblocks the corresponding delegate thread on the CPU. The delegate then executes the ROS 2 subscriber take function {\tt rcl\_take} on behalf of the hardware thread \circled{2}. When a message for the subscribed topic becomes available, the {\sc ReconROS} stack stores it in main memory \circled{3} and unblocks the delegate thread \circled{4}, which in turn sends the message pointer via the OSIF back to the hardware thread \circled{5}. Subsequently, the hardware thread can read the message via its MEMIF \circled{6}.

Publishing a message from a hardware thread works analogously: First, the hardware thread stores the message in the main memory. Then, it sends a {\tt ROS\_publish} command and the message pointer via the OSIF interface to its delegate thread, which executes the command.

The {\sc ReconROS} mechanism described in Figure~\ref{fig:reconros_take_sequence} also supports the implementation of 
ROS 2 services and actions. ROS 2 services comprise the receiving and sending of a single message, while the more involved
ROS 2 actions combine two ROS 2 services with a publish/subscribe feedback channel. %

\subsection{Design Flow}
\label{sec:ReconROS:designflow}

\begin{figure}[h]
	\center
    \includegraphics[width=\linewidth]{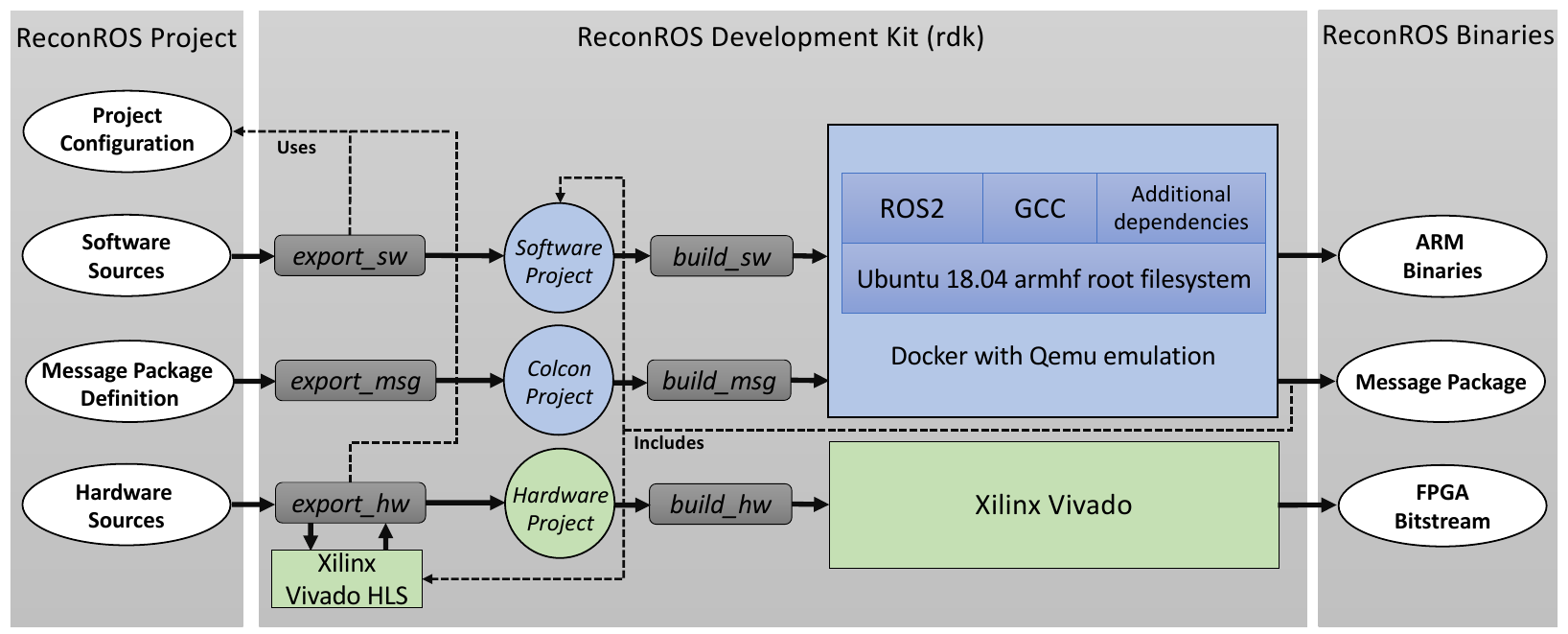}    
    \caption{ReconROS design flow}
    \label{fig:DesignFlow}
\end{figure}

The design flow for a {\sc ReconROS} application adapts the original ReconOS design flow~\cite{6636314} and is sketched in 
Figure~\ref{fig:DesignFlow}. The flow starts with the specification of a {\sc ReconROS} project comprising a project configuration file, the sources for software and hardware threads that represent the ROS 2 nodes, and the definition of message types used for the application.

The configuration file specifies the used ROS 2 objects with their dependencies, the ReconOS architecture including, in particular, the number of reconfigurable slots and the mapping of hardware threads to reconfigurable slots, and the settings for the build tool flow.

The basic element of each {\sc ReconROS} application is the \textit{rosnode} object, which represents a ROS 2 node in the network. A \textit{rosnode} object can be extended by one or more communication objects, which can be subscriber (\textit{rossub}) or publisher (\textit{rospub}) objects for specific topics in case of publish/subscribe communication, 
service (\textit{rossrvs} / \textit{rossrvc}) objects for client-server communications, and action (\textit{rosacts} / \textit{rosactc}) objects for ROS 2 actions. 
In addition, each of these extensions, i.e., publisher, subscriber, service, and action, requires a reference to an instance of a ROS message \textit{rosmsg} of a specific type. Declarations of \textit{rosmsg} objects include the communication type, a group, and the message type. For example, a specific message declaration could specific 'Image' as message type, 'sensor\_msgs' as group, and publish/subscribe as communication type.

Threads for ROS 2 software nodes can be developed in C and threads for ROS 2 hardware nodes in C/C++ for use with high-level synthesis or, alternatively in VHDL. Importantly, we provide the same {\sc ReconROS} API for software and hardware threads which greatly simplifies the creation of hardware-accelerated versions of software threads.

Based on the configuration file and the sources, the ReconOS development kit (rdk) creates the ReconROS binaries for the specific project.
The rdk command {\tt export\_msg} extracts information from the message package definition and creates a Colcon project, which is then compiled to the message package by the command {\tt build\_msg}. Colcon is a ROS 2 build tool, and the message package comprises message-related data and scripts that are used by the ROS 2 runtime. The rdk command {\tt export\_sw} creates the software project based on the sources for software threads and configuration data. The software project also includes the ReconOS delegate threads, all necessary initialization functions for the ReconOS primitives, and the ROS 2 middleware dependencies. Moreover, the software project includes header definitions for the messages, which are part of the compiled message package. Since we target Xilinx platform FPGAs of the Zynq-7000 series, which contain ARM Cortex-A9 cores, the rdk command {\tt build\_sw} creates binaries for the ARM architecture.

Both commands, {\tt build\_sw} and {\tt build\_msg} employ an ARM-32 docker container emulated with Qemu to build the binaries. Compared to a standard cross-compilation tool chain for the embedded ARM cores, our setup greatly simplifies the ROS 2 build step with all its dependencies since the package manager within the container can be used. 
Finally, the rdk command {\tt export\_hw} creates the hardware project based on the sources for hardware threads and configuration data. 
The hardware project contains the complete {\sc ReconROS} architecture with its OSIFs, MEMIFs, and supporting modules. The command calls Xilinx Vivado HLS for high-level synthesis and thus also requires the message header definitions. 
The FPGA bitstream is then created by the rdk command {\tt build\_hw}.

\subsection{Example ROS 2 Application}
\label{sec:ReconROS:Example}

\begin{figure}[!h]
	\center
    \includegraphics[width=0.62\linewidth]{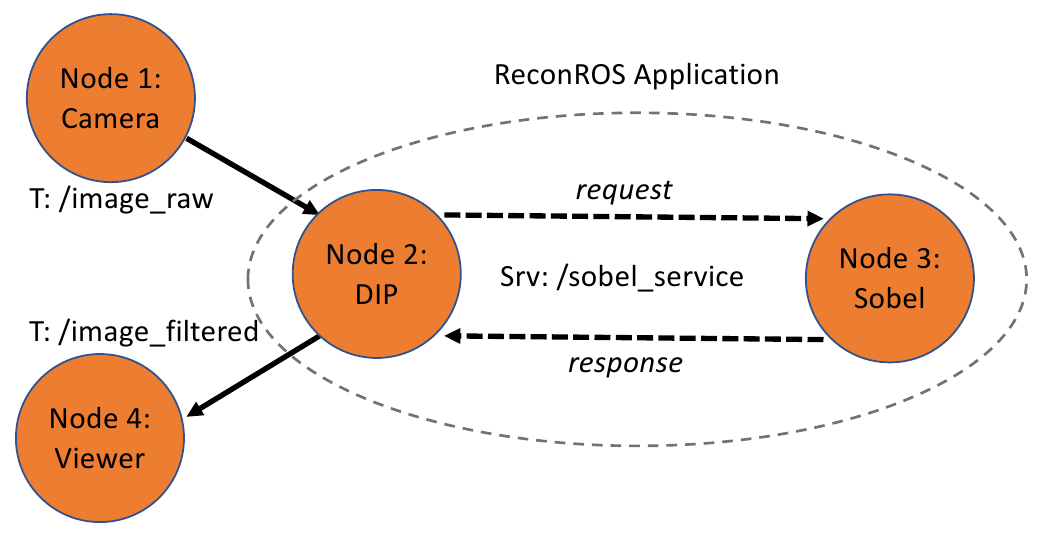}    
    \caption{Example ROS 2 application}
    \label{fig:example_ros_application}
\end{figure}

As an example we elaborate on a ROS 2 application comprising four nodes, which is shown in Figure~\ref{fig:example_ros_application}. Node 1 captures images from a camera and publishes them to the topic {\tt /image\_raw}. Node 2, the digital image processing node (DIP), subscribes to this topic, offloads the image processing to node 3, the Sobel filter node (Sobel), and publishes the filtered images to the topic {\tt /image\_filtered}. Node 4 reads and displays the filtered images. The data exchange between the Sobel and DIP nodes is done with a ROS 2 service called {\tt sobel\_service}. The ReconROS application comprises nodes 2 and 3, where both are to be mapped to reconfigurable hardware and run either on a single or on two FPGA platforms. Nodes 1 and 4 are assumed to be existing or being compiled with appropriate ROS 2 design flows to other target architectures, e.g., desktop PCs.

Listing~\ref{lst:configuration_file} shows the ROS 2-related part of the configuration file for the nodes 2 and 3.
The information for the ROS 2 nodes is organized into so-called resource groups. Lines 1--4 specify node 3, beginning with the definition of a rosnode object named "Sobel" in line 2. In line 3, a message object of type ROS 2 service message is defined with further references to a ROS 2 message package and the communication as well as service types. Line 4 declares a ROS 2 server object for a ROS 2 service, connects it to the ROS 2 node {\tt node\_3} and the message object {\tt filter\_service\_msg}, assigns the name "sobelservice" to it, and sets the polling time for checking for new service requests to $10000 \: \mu s$.

Lines 6--12 specify node 2, including the rosnode object named "DIP", the same message object as used by node 2, and a client object for a ROS 2 service. Additionally, node 2 is extended with the message object {\tt image\_msg} of a ROS 2 built-in message type and corresponding subscriber and publisher objects for the topics {\tt /image\_raw} and {\tt /image\_filtered}.

\noindent\begin{minipage}{\linewidth}
    \centering
\begin{lstlisting}[floatplacement=H, numbers=left, numbersep=-5pt, frame=single, caption={Configuration file (ROS 2-related part) for the 
    {\sc ReconROS} application shown in Figure~\ref{fig:example_ros_application}}\label{lst:configuration_file}] 
   [ResourceGroup(at)ResourceGroupSobel]
   node_3 = rosnode, "Sobel"
   filter_service_msg = rossrvmsg, application_msgs, srv, SobelSrv  
   filter_server = rossrvs, node_3, filter_service_msg, "sobelservice", 10000
 
   [ResourceGroup(at)ResourceGroupDIP]
   node_2 = rosnode, "DIP"
   filter_service_msg = rossrvmsg, application_msgs, srv, SobelSrv 
   filter_client = rossrvc, node_2, filter_service_msg, "sobelservice", 10000
   image_msg = rosmsg, sensor_msgs, msg, Image 
   sub = rossub, node_2, image_msg, "/image_raw", 10000
   pub = rospub, node_2, image_msg, "/image_filtered"     
\end{lstlisting}
\end{minipage}

Listing~\ref{lst:hls_implementation_sobel} presents C/C++ code for the HLS-implementation of the "Sobel" ROS 2 node. Using the {\sc ReconROS} API, the processing loop starts in line 3 with a blocking read for a new service request. When a request becomes available, the function {\tt ROS\_SERVICESERVER\_TAKE} returns a pointer to the service request data structure. With the help of the {\tt OFFSETOF} macro, line 4 determines another pointer to the address of the request's payload. The macro {\tt MEM\_READ} is employed to first read the address of the image 
in line 7 and then to read the image into a {\tt ram} structure within the FPGA in line 8. After a Sobel filter function is executed on the image in line 10, the result is written back to main memory via the {\tt MEM\_WRITE} macro. Finally, the node sends the filtered data back to the node requesting the filter service. ({\tt ROS\_SERVICESERVER\_SEND\_RESPONSE}). This code example shows the steps required to create a ReconROS application and, hence,  focuses on simplicity rather than on optimized performance. For example, overlapping processing with memory transfers using a line buffer approach would be a natural optimization.

\noindent\begin{minipage}{\linewidth}
    \begin{lstlisting}[floatplacement=H,numbers=left, numberfirstline=true, numbersep=-5pt, frame=single,caption={C/C++ code (partial) for the HLS implementation of the "Sobel" ROS 2  node}\label{lst:hls_implementation_sobel}] 
  while(1) {   
   // Wait for service request and get pointer to payload
   pMsg = ROS_SERVICESERVER_TAKE(resourcedip_srv, resourcedip_filter_srv_req);
   pMsg += OFFSETOF(application_msgs__srv__SobelSrv_Request, img.data.data);

   // Get pointer to image in memory and copy it to FPGA-internal memory
   MEM_READ(pMsg, pPayloadService, 4);  
   MEM_READ(pPayloadService[0], ram, IMAGE_SIZE * 4); 

   SobelFilter(ram);

   // Write filtered image back to memory and send service response
   MEM_WRITE(ram, pPayloadService[0], IMAGE_SIZE * 4); 
   ROS_SERVICESERVER_SEND_RESPONSE(resourcedip_srv, resourcedip_filter_srv_res);
  }
\end{lstlisting}
\end{minipage}

Listing~\ref{lst:hls_implementation_dip} displays a similar procedure for the "DIP" node, which is expanded with three communication objects, a subscriber object for the topic {\tt /image\_raw}, a client object for the service {\tt /sobel\_service}, and a publisher object for the topic {\tt /image\_filtered}.

\noindent\begin{minipage}{\linewidth}
\begin{lstlisting}[floatplacement=H,numbers=left, numbersep=-5pt, frame=single,caption={C/C++ code (partial) for the HLS implementation of the "DIP" ROS 2 node}\label{lst:hls_implementation_dip}] 
  while(1) {   
   // Wait for published image and get pointer to payload
   pMsg = ROS_SUBSCRIBER_TAKE(resourcesobel_subdata, resourcesobel_image_msg);
   pMsg += OFFSETOF(sensor_msgs__msg__Image, data.data);
    
   // Get pointer to image in memory and copy it to FPGA-internal memory
   MEM_READ(pMsg, pPayloadPubSub, 4);
   MEM_READ(pPayloadPubSub[0], ram, IMAGE_SIZE * 4);

   // Request filter service, pServiceRequest is set up during initialization
   MEM_WRITE(ram, pServiceRequest[0], IMAGE_SIZE * 4);
   ROS_SERVICECLIENT_SEND_REQUEST(resourcesobel_srv,resourcesobel_filter_srv_req);

   // Wait for service response and get pointer to payload 
   pMsg = ROS_SERVICECLIENT_TAKE(resourcesobel_srv, resourcesobel_filter_srv_res);
   pMsg += OFFSETOF(application_msgs__srv__SobelSrv_Response, img.data.data);
    
   // Get pointer to payload and copy it to FPGA-internal memory
   MEM_READ(pMsg, pPayloadService, 4);
   MEM_READ(pPayloadService[0], ram, IMAGE_SIZE * 4);

   // Write filtered image back to memory and publish it
   MEM_WRITE(ram, pPayloadPubSub[0], IMAGE_SIZE * 4);
   ROS_PUBLISHER_PUBLISH(resourcesobel_pubdata, resourcesobel_image_msg);
  }
\end{lstlisting}
\end{minipage}

\section{Evaluation}
\label{sec:Evaluation}

In this section, we first describe experiments to quantify the overheads involved for mapping ROS 2 nodes to hardware, followed by a distributed mechatronics application example that demonstrates the feasibility and flexibility of {\sc ReconROS}.

\begin{figure}[ht]
	\center
    \includegraphics[width=0.7\linewidth]{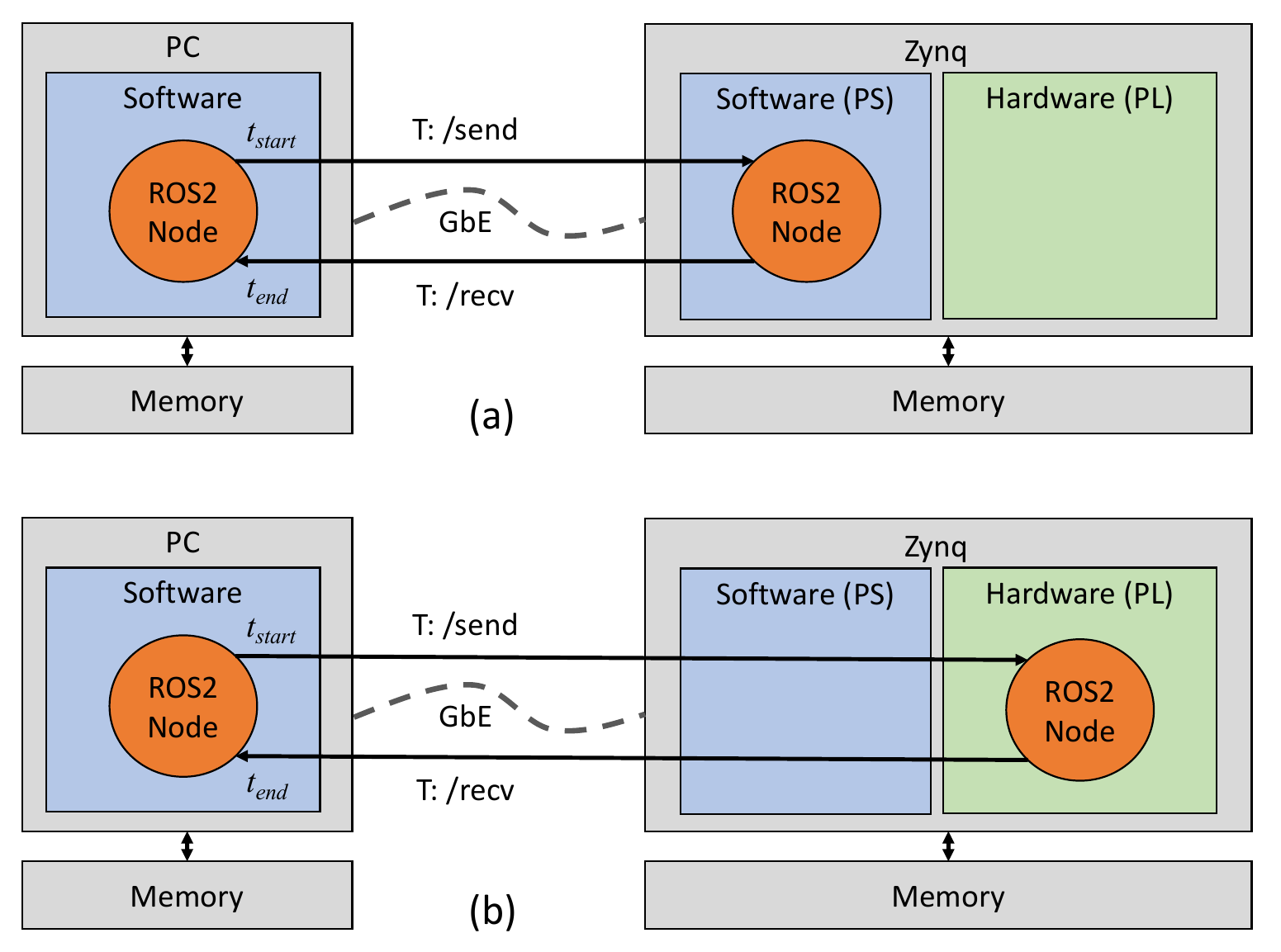}    
    \caption{{\sc ReconROS} ping-pong application}
    \label{fig:reconros_experiments_roundtrip}
\end{figure}

\subsection{ROS 2 Hardware Node Overheads}
\label{sec:Experiments:latency}

To characterize runtime overheads when mapping ROS 2 nodes to hardware instead of software, and contrasting them to communication times within a ROS 2 network, we have implemented a ping-pong {\sc ReconROS} application with two ROS 2 nodes distributed onto a desktop PC and a Mini-ITX 7Z100 board containing a Xilinx Zynq-7100 platform FPGA, connected via Gigabit Ethernet (GbE) as shown in Figure~\ref{fig:reconros_experiments_roundtrip}. The platform FPGA runs Ubuntu 18.04 and {\sc ReconROS} based on ROS 2 dashing. All ROS 2 nodes use the same C/C++ source for software and hardware implementations. Software implementations have been compiled with optimizations level O3, and hardware implementations have been created with HLS without any optimizations. All reported runtimes have been averaged over 1000 executions. 

The first experiment determines the basic overhead for mapping a ROS 2 node to hardware and consists of an echo application, where the ROS 2 node on the PC publishes messages to the topic {\tt T:/send} and the ROS 2 node on the Zynq subscribes to this topic, receives messages in local memory and publishes them to the topic {\tt T:/recv}. Table~\ref{table:Experiments:latency:echo} presents the runtimes for the echo tasks in software and hardware, $t_{pp-echo-SW}$ and $t_{pp-echo-HW}$, measured as $t_{pp}=t_{end}-t_{start}$ on the PC as indicated in Figure~\ref{fig:reconros_experiments_roundtrip}, and the resulting speedup $S_{pp-echo}$.

\begin{table}[ht]
    \begin{center}
        \begin{tabular}{|l | r r r |}
        \hline
        \makecell[c]{Message \\ {size}}  & \makecell{$t_{pp-echo-SW}$ \\ {[}ms{]}}  & \makecell{$t_{pp-echo-HW}$ \\ {[}ms{]}} & \makecell{$S_{pp-echo}$ \\ {}} \\[0.5ex] 
        \hline 
        \makecell[c]{4 Byte}   &  0.81   &  1.2    & 0.68      \\
        \hline
        \makecell[c]{8 KiB}    & 10.65   & 10.48   & 1.02      \\
        \hline
        \makecell[c]{1 MiB}    & 52.21   & 52.16   & 1.01      \\
        \hline
        \makecell[c]{6 MiB}    & 363.91  & 363.30  & 1.00   \\
        \hline
        \makecell[c]{10 MiB}    & 630.37   & 624.02  & 1.01  \\        
        \hline
        \end{tabular}
     \end{center}
    \caption{Runtimes and speedups for the echo ping-pong application}
    \label{table:Experiments:latency:echo}
\end{table}

The echo software node does actually not perform any operations except calling subscribe/publish functions. To implement the same behavior, the echo hardware node needs ReconOS signaling to communicate between the underlying hardware thread and the software-bound delegate thread. Since only pointers to messages and identifiers for the topics and the message are passed, the echo nodes exhibit a runtime independent of the message size. As Table~\ref{table:Experiments:latency:echo} shows, for the very small message size of 4 Byte there is a measurable slowdown due to the ReconOS signaling, but for larger message sizes this overhead is completely hidden behind the communication times. It has to be noted that mapping a ROS 2 node to hardware reduces the load on the CPU and this can become a source for additional speedups for the overall ROS 2 applications. Such an effect, albeit very small, can be observed in the echo experiment where some speedups are slightly larger than one.

The second experiment is a copy application that evaluates the memory read/write performance of ROS 2 hardware nodes. The difference to the echo application is that the Zynq-bound ROS 2 nodes create a copy of the message in local memory before publishing to topic {\tt T:/recv}. Table~\ref{table:Experiments:latency:copy} presents the runtimes for the raw copy tasks in software and hardware, $t_{raw-copy-SW}$ and $t_{raw-copy-HW}$, and the resulting raw speedup $S_{raw-copy}$, as well as the runtimes for the overall copy ping-pong application, $t_{pp-copy-SW}$, $t_{pp-copy-HW}$, and the resulting speedup $S_{pp-copy}$ for different message sizes.

\begin{table}[ht]
    \begin{center}
        \begin{tabular}{|l | r r r | r r r|}
        \hline
        \makecell[c]{Message \\ {size}}  & \makecell{$t_{raw-copy-SW}$ \\ {[}ms{]}}  & \makecell{$t_{raw-copy-HW}$ \\ {[}ms{]}} & \makecell{$S_{raw-copy}$ \\ {}} & \makecell{$t_{pp-copy-SW}$ \\ {[}ms{]}}     &  \makecell{$t_{pp-copy-HW}$ \\ {[}ms{]}}   & \makecell{$S_{pp-copy}$ \\ {}}  \\[0.5ex] 
        \hline 
        \makecell[c]{4 Byte}   &  0.01   & 0.01   & 1.00 & 1.69    & 1.71     &  0.99    \\
        \hline
        \makecell[c]{8 KiB}    & 0.03    & 0.13   & 0.23 & 11.39   & 10.78    &  1.06    \\
        \hline
        \makecell[c]{1 MiB}    & 3.59    & 12.81  & 0.28 & 58.71   & 66.25    &  0.89     \\
        \hline
        \makecell[c]{6 MiB}    & 18.91   & 76.35  & 0.25 & 381.44  & 438.03   &  0.87   \\
        \hline
        \makecell[c]{10 MiB}   & 31.54   & 127.19  & 0.25 & 643.47  & 735.30   &  0.86   \\        
        \hline
        \end{tabular}
     \end{center}
    \caption{Runtimes for the raw copy ROS 2 nodes in software and hardware and for the overall copy ping-pong application, and corresponding speedups}
    \label{table:Experiments:latency:copy}
\end{table}

Since the underlying ReconOS implementation has a lower memory bandwidth compared to the Zynq's ARM processor subsystem, we observe a slowdown for the raw ROS 2 hardware copy node, which is distinct for larger message sizes and saturates at about $0.25$. Thus, copying a message of 10 MiB is about $4 \times$ slower in hardware than in software. While improving ReconOS' memory subsystem would obviously improve the situation, Table~\ref{table:Experiments:latency:copy} also shows that for the overall copy ping-pong application where we have to take communication into account the slowdown is less pronounced and saturates at around $0.86$. Again, due to effects of the underlying software stacks of Linux, ROS 2, and ReconOS, and the possible parallel execution of hardware and software threads, the speedups are not consistently decreasing and for 8 KiB the speedup is even larger than one.

Related work~\cite{Sugata2017} has also reported on measured communication times between a ROS node on a PC and a ROS software node on an ARM/Zynq connected with Gigabit Ethernet. For a one-way communication the authors determined approximately 60 ms for a 1 MiB message and approximately 275 ms for a 6 MiB message. Comparing with the corresponding data points of Table~\ref{table:Experiments:latency:echo}, which are for two-way communication, we see that {\sc ReconROS} achieves higher performance, albeit  on a different ROS version. 

In a third experiment, we have have implemented the following four smaller applications on the platform shown in Figure~\ref{fig:reconros_experiments_roundtrip}:

\paragraph{Inverse kinematics:}
This application computes control signals for driving a servo motor that sets a joint angle based on a desired position and orientation of a robotic manipulation platform. The application is part of a larger mechatronic system~\cite{Lienen_2019} for controlling the movements of a Stewart platform~\cite{stewartplatform} with six degrees of freedom. The computation involves coordinate transformations and an iterative implementation of the $\arctan()$ function. The ROS 2 input message is an unsigned 32 Bit integer packed with two fixed-point numbers in Q8.6 format that represent the desired rotation angles of the platform around the x-axis and the y-axis. The ROS 2 output messages is also a 32 Bit unsigned integer containing a 10 Bit unsigned integer which is the pulse width coded control signal for the motor.  

\paragraph{Number sorting:}
This application sorts an array of 32 Bit unsigned integers based on the odd-even transposition sort algorithm~\cite{knuth1998art}. The algorithm is based on a comparator network that employs $n$ stages with $n$ comparisons each to sort $n$ numbers. The ROS 2 node on the PC generates random numbers and publishes messages comprising 2048 numbers as an array. The Zynq-based ROS 2 node sorts the data and sends it back. 

\paragraph{Sobel filter:} 
This application implements a Sobel image filter~\cite{gonzalez2018digital} operating on three channels (RGB) of dimension $640 \times 480$. The filter applies two filter kernels on each channel of the image and calculates the absolute value of the dot product as an approximation for the geometric mean. The ROS 2 input and output messages are of the type {\tt Image} from the ROS 2 sensor message package.

\paragraph{MNIST classifier:} 
This application classifies handwritten digits from the MNIST dataset by implementing a neural network. The classifier is implemented as a ROS 2 service, which accepts input request images of size $28 \times 28$ as custom ROS 2 messages and response the estimated digit. The classifier consists of three convolution layers, three pooling layers and two fully connected layers. The achieved accuracy is about 97\%.

\begin{table}[ht]
     \begin{center}
         \begin{tabular}{|l r r r|}
         \hline
         \makecell[l]{{\sc ReconROS} application}        & \makecell{Slice LUTs}    & \makecell{DSP}    &   \makecell{BRAM}   \\ [0.5ex]
         \hline
         \makecell[l]{Inverse kinematics}                & 4802 (1.73\%)            & 17 (0.84\%)       &   3 (0.40\%)   \\
         
         \makecell[l]{Number sorting}                    & 10396 (3.75\%)           & 0 (0.00\%)        &   2 (0.26\%) \\

         \makecell[l]{Sobel filter }                     & 13625 (4.91\%)           & 0 (0.00\%)        &   10 (1.32\%)  \\

         \makecell[l]{MNIST classifier}                  & 26071 (9.40\%)           & 18 (0.89\%)      &  57.5 (7.62\%)\\
         \hline
         \end{tabular}
      \end{center}
      \caption{Resource usage and utilization (in \% of the Xilinx Zynq 7100) for the implemented {\sc ReconROS} applications. 
	  Resource figures are reported for look-up tables (Slice LUTs), digital signal processing blocks (DSP), and block memory (BRAM)}
      \label{table:Experiments:resource_usage}
\end{table}

Table~\ref{table:Experiments:resource_usage} displays resource usage and FPGA utilization for the four applications, and Table~\ref{table:Experiments:runtime_comparison} the raw runtimes for the Zynq-bound ROS 2 nodes, which are either mapped to the ARM core ($t_{raw-SW}$) or to reconfigurable logic ($t_{raw-HW}$), the resulting raw speedup $S_{raw}$, as well as the runtimes for the overall applications measured in the ping-pong fashion shown in Figure~\ref{fig:reconros_experiments_roundtrip}. 

The inverse kinematics application achieves a raw ROS 2 node speedup of $6.32 \times$, the sobel filter and MNIST classifier also achieve raw speedups, but the number sorting application does not benefit from hardware mapping. It has to be noted that the goal of these experiments has been to evaluate the overheads involved for {\sc ReconROS} applications rather than achieving high speedups through hardware acceleration. There is obviously potential to improve the raw speedups for the hardware-mapped ROS 2 nodes, in particular for the number sorting application where more parallelism can be exploited. Depending on the relation between communication and computation times, the speedups for the overall ROS 2 applications are sometimes considerably lower than the raw speedups, i.e., for inverse kinematics, sometimes slightly lower, i.e., for the sobel filter and the MNIST classifier, and in the case of number sorting even slightly higher.

\begin{table}[ht]
    \begin{center}
        \begin{tabular}{|l | r | r  r  r  | r  r  r|} 
        \hline
        \makecell[l]{{\sc ReconROS}\\application}  & \makecell{ Message Size \\ In/Out } & \makecell{$t_{raw-SW}$ \\ {[}ms{]}}  & \makecell{$t_{raw-HW}$ \\ {[}ms{]}}   & \makecell{$S_{raw}$ \\ {}}  & \makecell{$t_{pp-SW}$ \\  {[}ms{]}}  & \makecell{$t_{pp-HW}$ \\ {[}ms{]}}  & \makecell{$S_{pp}$ \\ {}} \\ 
        \hline
        \makecell[l]{Inverse }          & 4/4 Byte   & 1.20       & 0.19       &   6.32      &   7.70       &  6.64    &   1.16 \\
		
        \makecell[l]{Sorting }          & 8/8 KiB      & 17.44      & 35.11      &   0.50      &   24.42      & 42.08    &   0.58 \\ 
	   	
        \makecell[l]{Sobel }            & 900/900 KiB  & 37.53      & 22.28      &   1.68      &   83.39      & 68.54    &   1.22 \\ 
        
        \makecell[l]{MNIST }            & 850/4 Byte   & 88.03      & 30.74      &   2.86      &   98.58      & 41.25    &   2.39 \\
        \hline
       \end{tabular}
     \end{center}
    \caption{Runtimes of software and hardware ROS 2 nodes and for the overall applications, and corresponding speedups. The hardware implementations are not optimized for performance}
    \label{table:Experiments:runtime_comparison}
\end{table}

To summarize the set of experiments detailed in this section: We have shown that while there is an overhead for mapping a ROS 2 node to hardware, the impact on an overall ROS 2 application depends on many factors such as i) the raw speedup of the ROS 2 hardware node, ii) the message size, iii) the overall application's topology and involved communication patterns and times, and iv) the ratio between node computation times and communication times. The memory access performance for ROS 2 hardware nodes is lower than for their software counterparts, an aspect that will be addressed as part of future work. Additional speedups can be realized through the parallel execution of hardware and software threads.

Finally, all hardware and software versions of the {\sc ReconROS} applications are semantically identical. Creating the different versions simply requires a change in the {\sc ReconROS} configuration file before running the functions of the {\sc ReconROS} development kit. This flexibility in generating variants of ROS 2 hardware-accelerated applications is the main feature of  {\sc ReconROS}.

\subsection{Mechatronics Model}

\begin{figure}[ht]
	\center
    \includegraphics[width=0.9\linewidth]{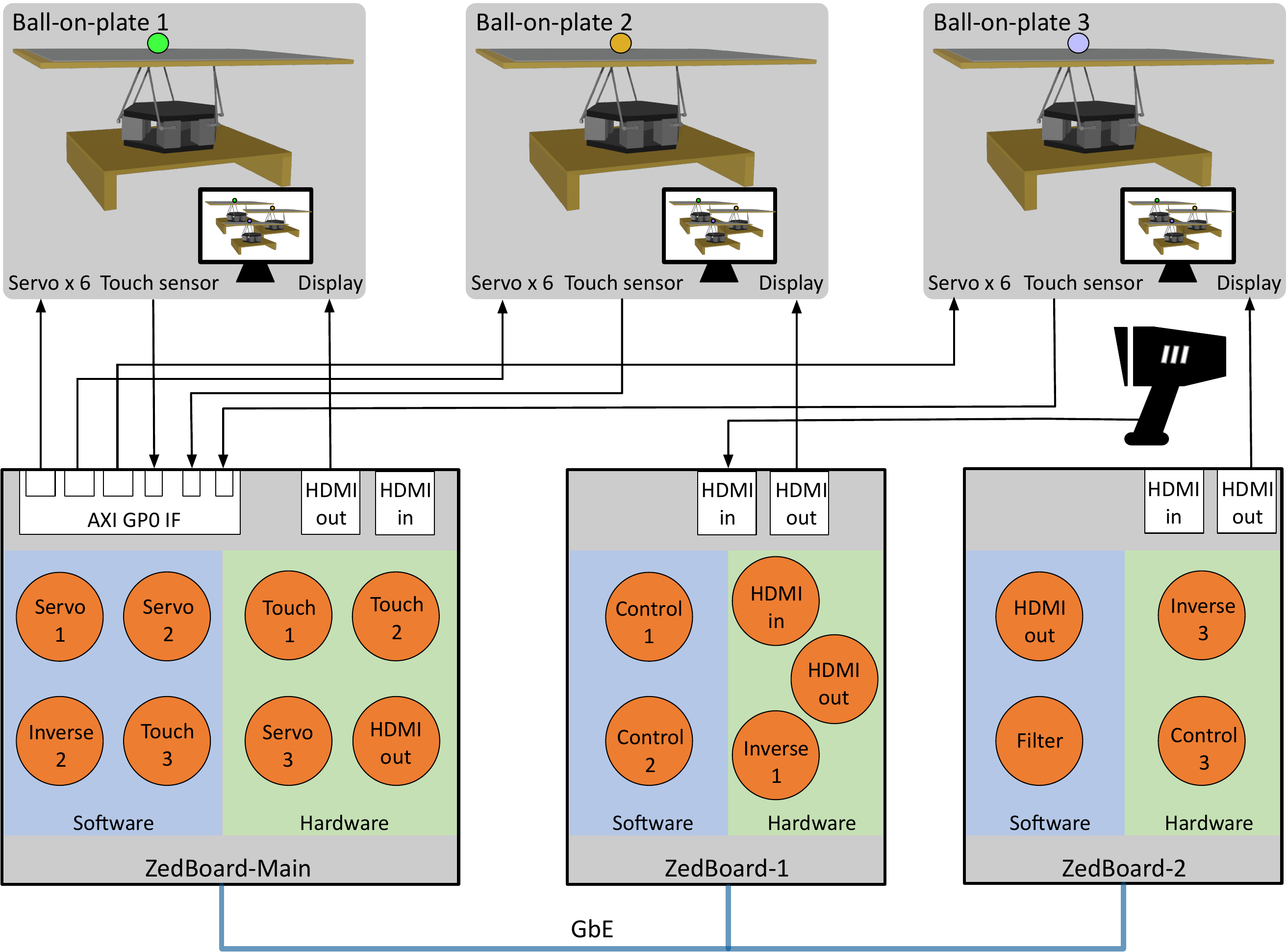}    
    \caption{Mechatronics model based on three ball-on-plate stations with Stewart platforms}
    \label{fig:bop_construction}
\end{figure}

To showcase the suitability of {\sc ReconROS} for distributed hardware-accelerated ROS 2 applications we present the mechatronics model~\cite{Lienen_2019} shown in Figure~\ref{fig:bop_construction}, that we have physically implemented.
The model comprises three ball-on-plate stations that are able to 
balance a mechanical platform such that a ball thrown onto the platform does not fall off. To this end we employ a Stewart platform \cite{stewartplatform} that allows the system to move an object in six degrees of freedom, including linear translations in $x$, $y$ and $z$ direction but also three rotations (pitch, roll, and yaw). Stewart platforms are perfectly suitable for high dynamic mechatronics  applications such as flight simulators or telescopes. In our setup, we drive six servo motors by pulse-width modulated signals to adjust 
corresponding angles between the motor axes and the legs connecting to the platform, which then results in the wanted movement. To capture the position $(x,y)$ of the ball on the platform we use a resistive touchscreen mounted on the surface of the platform. 
Additionally, each ball-on-plate station is equipped with a monitor, and a camera is capturing all stations.

The computing infrastructure includes three ZedBoards, as outlined in Figure~\ref{fig:bop_construction}. Each ZedBoard is equipped with a Xilinx Zynq-7020 platform FPGA and runs Ubuntu 18.04 and {\sc ReconROS} based on ROS 2 dashing. The servo actuators and touchscreen sensors are connected to ZedBoard-Main, the camera is connected to ZedBoard-1 and the monitor inputs on the three ball-on-plate stations are driven by a ZedBoard each. All compute platforms are connected in an Ethernet network.

\begin{figure}[ht]
	\center
    \includegraphics[width=0.85\linewidth]{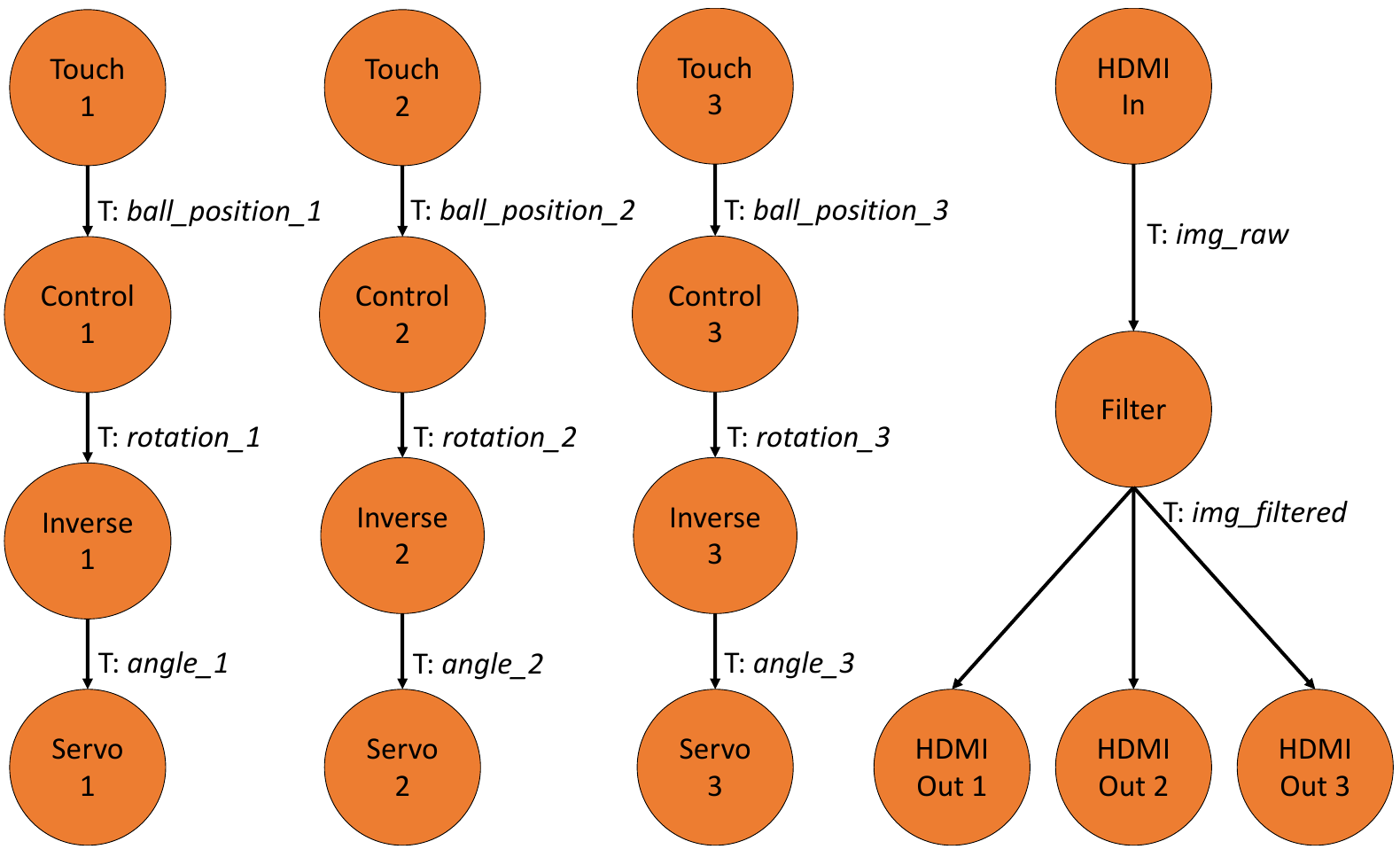}    
    \caption{ROS 2 application with node and communication objects for the mechatronics model shown in Figure~\ref{fig:bop_construction} }
    \label{fig:bop_sw_architecture}
\end{figure}

The overall ROS 2 application splits into two parts, the control of ball-on-plate stations and a video processing chain. Figure~\ref{fig:bop_sw_architecture} shows all involved ROS 2 nodes with their communication objects. The control loop for a ball-on-plate station comprises the four ROS 2 nodes {\tt Touch}, {\tt Control}, {\tt Inverse} and {\tt Servo}. The {\tt Touch} node starts a new control cycle by reading the actual position of the ball on the platform. This information is scaled and sent to the {\tt Control} node that implements a PID controller and a Kalman filter to determine the desired rotations for the platform with respect to the $x$ and $y$ axes. The subsequent {\tt Inverse} node applies inverse kinematics transformations to determine the required angle for each of the six servo motors. Finally, the {\tt Servo} node converts the angles into pulse width modulated signals to drive the motors. 

The video processing chain includes ROS 2 nodes for video input, {\tt HDMI in}, processing, {\tt Filter}, and video output, {\tt HDMI out}. The HDMI interface implementation includes mechanisms for the transport of image data from and to the main memory without processor interaction by using AXI VDMAs (Video Direct Memory Access).
All ROS 2 nodes use publish/subscribe mechanisms to communicate with topics shown in Figure~\ref{fig:bop_sw_architecture}.

We have realized all ROS 2 nodes in software and hardware. Table~\ref{tab:Exectimes_bop} lists the raw node runtimes. The hardware implementations of the inverse kinematics and the filter nodes can exploit low-level parallelism and achieve speedups. All other nodes are either more control-flow intensive, exhibit little computation, or are bound by the memory bandwidth and are thus better mapped to software. 

Given that both software and hardware implementations for the ROS 2 nodes are available, developers can easily distribute the nodes across the boards in the network, change the mapping of nodes in the project configuration files, and re-build the system. One specific example for such a mapping of nodes is indicated in Figure~\ref{fig:bop_construction}. With this mapping, the sampling time of the {\tt Touch} node and, thus, the control loop could be set to 20 ms which results in rather smooth movements of the Stewart platforms. Table~\ref{table:bop:resource_usage} lists the resources required for this specific mapping, including the actual hardware-mapped nodes, the necessary {\sc ReconROS} infrastructure, and the components needed for the HDMI input and output interfaces.

\begin{table}[ht]
    \begin{center}
        \begin{tabular}{|l | r  r r |} 
        \hline
        \makecell[l]{ROS 2 node}  & \makecell{$t_{raw-node-SW}$ \\ {[}ms{]}}  & \makecell{$t_{raw-node-HW}$ \\ {[}ms{]}}  & \makecell{$S_{raw-node}$ \\ {}} \\ 
        \hline
        \makecell[l]{Servo}          & 0.001       & $<$ 0.001      &  $\approx$1 \\		
        \makecell[l]{Control}        & 0.017       & 0.030          &  0.57 \\    	
        \makecell[l]{Inverse}        & 1.430       & 0.196          &  7.30 \\        
        \makecell[l]{Touch}          & 0.001       & $<$ 0.001      & $\approx$1 \\
        \hline
		\makecell[l]{HDMI In}        & 5.160       & 18.460         &  0.28 \\
		\makecell[l]{HDMI Out}       & 4.590       & 18.400         &  0.25 \\
		\makecell[l]{Filter}         & 37.530      & 22.280         &  1.68 \\	
		\hline
       \end{tabular}
     \end{center}
    \caption{Runtimes for the raw ROS 2 nodes of the mechatronics example in software and hardware}
    \label{tab:Exectimes_bop}
\end{table}

\begin{table}[ht]
    \begin{center}
        \begin{tabular}{|l r r r r|}
        \hline  
        \makecell[l]{Board}        &  \makecell{FPGA}   & \makecell{Slice LUTs}    & \makecell{DSP}     &   \makecell{BRAM}   \\ [0.5ex]
        \hline
        \makecell[l]{Zedboard Main} &   Zynq-7020       & 13467 (25.31\%)           & 0 (0.00\%)        &   3 (2.14\%)   \\
        
        \makecell[l]{Zedboard 1}    &   Zynq-7020       & 13235 (24.88\%)           & 77 (35.00\%)     &    3 (2.14\%) \\

        \makecell[l]{Zedboard 2}    &   Zynq-7020       & 13031 (24.49\%)           & 77 (35.00\%)     &    3 (2.14\%) \\

        \hline
        \end{tabular}
     \end{center}
     \caption{Resource usage and utilization (in \% of the Xilinx Zynq 7020) for the three involved FPGA-boards. Resource figures are reported for look-up tables (Slice LUTs), digital signal processing blocks (DSP), and block memory (BRAM)}
     \label{table:bop:resource_usage}
\end{table}

We want to note that the implemented system is not a hard real-time system with a guaranteed sampling period of 20 ms. Creating a hard real-time system would require to modify ReconROS and the underlying ROS 2 and Linux layers, as well as substitute Ethernet communication with a real-time version and is clearly out of scope for this work.

\begin{figure}[ht]
	\center
    \includegraphics[width=0.95\linewidth]{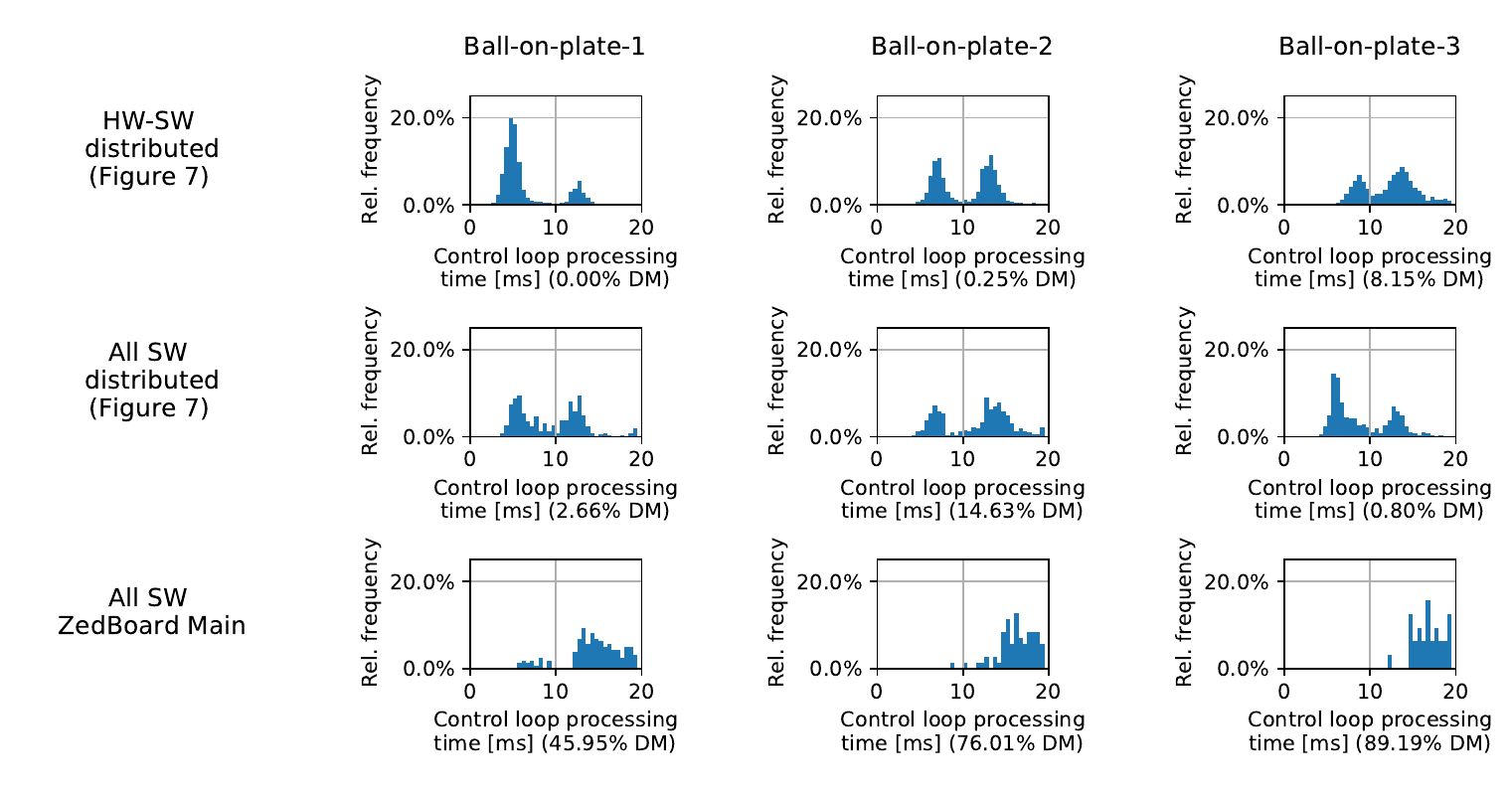}    
    \caption{Relative frequencies of measured processing times for the three control paths (Touch$\rightarrow$Control$\rightarrow$Inverse$\rightarrow$Servo) 
	and three different node mappings. (DM~=~Deadline~Missed~[\%])}
    \label{fig:bop_cycletimes}
\end{figure}

Moreover, the optimization of the mapping of nodes between hardware and software and across the FPGA board is also not addressed in this work. However, to demonstrate the trade-offs involved we have created three mappings of the mechatronics application and measured the processing times of the three control loops for 300 sampling periods. Figure~\ref{fig:bop_cycletimes} displays the relative frequencies for the resulting processing times for all three control loops, i.e., for the three Stewart platforms, (columns 1-3) and for three different ROS 2 mappings (rows 1-3). The figure shows the processing time frequencies from 0 to 20 ms and additionally provides the percentage of missed deadlines, where the deadline has been set to 20 ms.  

The first row uses the mapping from Figure~\ref{fig:bop_construction}, that distributes the nodes over all three FPGA boards and over software and hardware.  This mapping reaches all deadlines for platform 1, misses 0.25\% of the deadlines for platform 2, and 8.15\% for platform 3. The second row shows the same distribution of nodes across the three FPGA boards but maps all nodes to software. In this case, the fraction of missed deadlines is rather low on platforms 1 and 3, and with 14.63\% somewhat higher on platform 2. Finally, the mapping of row 3 places all nodes in software on Zedboard-Main with the result that most of the deadlines are missed. The Stewart platforms for mappings 1 and 2 move rather smoothly, but for mapping 3 the platforms show very jerky movements making the application unusable.

\section{Conclusion and Future Work}
\label{sec:ConclusionFuture Work}

In this paper we have presented {\sc ReconROS}, a novel approach that enables developers of ROS 2 robotics applications to leverage the performance and energy-efficiency of FPGA implementations. {\sc ReconROS} bases on ReconOS and allows for flexible hardware acceleration of ROS 2 nodes through an API that supports a consistent programming model for ROS 2 nodes across the hardware/software boundary, while preserving the main advantages of ReconOS such as full memory access for hardware threads or operating system like synchronization mechanisms for hardware/software co-designed applications.

Future work is planned along the following lines: First, we want to leverage partial reconfiguration available with ReconOS~\cite{6636314} to manage the reprogrammable hardware resources more efficiently, for example by configuring ROS 2 hardware nodes on demand. Second, since in distributed ROS networks not all compute nodes might be equipped with platform FPGAs, we plan to investigate the feasibility of a ROS 2 node offering acceleration-as-a-service. Third, while programming distributed robotics applications with FPGA acceleration is greatly supported by {\sc ReconROS}, there is a demand for simulating such systems before deployment and for adding runtime monitoring functionality that can be used for debugging. Finally, we plan to showcase {\sc ReconROS} for multi-drones~\cite{SchRin2020} which are one of the most demanding classes of distributed robotics systems.

\bibliographystyle{ACM-Reference-Format}
\bibliography{lienen21_acmtrets}

\end{document}